\documentclass[10pt, twocolumn]{article}
\usepackage{authblk}
\usepackage[utf8]{inputenc}
\usepackage{CJKutf8}
\usepackage{xcolor} 
\usepackage{mathtools}
\usepackage{ulem}
\usepackage[pdftex]{graphicx}
\usepackage{amsmath}
\usepackage{amssymb}
\usepackage{dsfont}
\usepackage{wrapfig}
\usepackage[top=25truemm,bottom=20truemm,left=20truemm,right=20truemm]{geometry}

\newcommand{\ja}[1]{}

\DeclarePairedDelimiter{\rbra}{\lparen}{\rparen} 
\DeclarePairedDelimiter{\cbra}{\lbrace}{\rbrace}
\DeclarePairedDelimiter{\sbra}{\lbrack}{\rbrack}

\DeclarePairedDelimiterX{\rset}[2]{\lparen}{\rparen}{#1\,\delimsize\vert\,#2}
\DeclarePairedDelimiterX{\set}[2]{\lbrace}{\rbrace}{#1\,\delimsize\vert\,#2}

\DeclarePairedDelimiterXPP{\prob}[2]{#1}{\lparen}{\rparen}{}{#2}
\DeclarePairedDelimiterXPP{\cond}[3]{#1}{\lparen}{\rparen}{}{#2\,\delimsize\vert\,#3}
\DeclarePairedDelimiterXPP{\KL}[2]{\mathrm{KL}}{\lbrack}{\rbrack}{}{#1\,\delimsize\|\,#2}
\DeclarePairedDelimiterXPP{\E}[2]{\mathds{E}}{\lbrack}{\rbrack}{_{#1}}{#2}

\makeatletter
\renewcommand{\AB@affilsep}{\quad\protect\Affilfont}

\makeatother

\title{\textbf{Goal-oriented inference of environment\\ from redundant observations}}
\author[1]{Kazuki Takahashi}
\author[2]{Tomoki Fukai}
\author[3]{Yutaka Sakai}
\author[1]{Takashi Takekawa}
\affil[1]{Informatics Program, Graduate School of Engineering, Kogakuin University of Technology and Engineering, Japan.}
\affil[2]{Neural Coding and Brain Computing Unit, Okinawa Institute of Science and Technology, Japan.}
\affil[3]{Brain Science Institute, Tamagawa University, Japan}

\date{}

\begin{document}

\twocolumn[
  \begin{@twocolumnfalse}
    \maketitle
    \begin{abstract}
      The agent learns to organize decision behavior to achieve a behavioral goal, such as reward maximization, and reinforcement learning is often used for this optimization. Learning an optimal behavioral strategy is difficult under the uncertainty that events necessary for learning are only partially observable, called as Partially Observable Markov Decision Process (POMDP). However, the real-world environment also gives many events irrelevant to reward delivery and an optimal behavioral strategy. The conventional methods in POMDP, which attempt to infer transition rules among the entire observations, including irrelevant states, are ineffective in such an environment. Supposing Redundantly Observable Markov Decision Process (ROMDP), here we propose a method for goal-oriented reinforcement learning to efficiently learn state transition rules among reward-related "core states'' from redundant observations. Starting with a small number of initial core states, our model gradually adds new core states to the transition diagram until it achieves an optimal behavioral strategy consistent with the Bellman equation. We demonstrate that the resultant inference model outperforms the conventional method for POMDP. We emphasize that our model only containing the core states has high explainability. Furthermore, the proposed method suits online learning as it suppresses memory consumption and improves learning speed.
    \end{abstract}
  \end{@twocolumnfalse}
]

\section{Introduction}
To accomplish a behavioral goal, agents should accurately perceive an environment through observations and generate adequate actions. Reinforcement learning (RL) is a paradigm in which an agent learns behavioral strategies to generate optimal actions appropriate to the given situation through trial and error. Actions that are likely to attain a behavioral goal are rewarded, so the strategy that maximizes the sum of long-term rewards is optimal. Recently, RL integrated with deep neural networks predicting the long-term reward has achieved high performance that exceeds human performance in complex applications such as the Atari game \cite{mnih2013playing, DBLP:journals/corr/HausknechtS15}. This further demonstrates the importance of the learning paradigm in real-world applications. 

When the environment changes dynamically, RL faces difficulty because the variety of observations increases. The true state space of the real-world environment generally consists of a relatively small number of task-relevant states and many other task-irrelevant states, where only the former is relevant to the reward. It is difficult for RL to follow the environment changes when the reward-relevant subset of states, which we may term core states, is entirely unknown. To improve the applicability of RL to real-world problems, a theoretical framework is needed to self-organize the reward-related state space from redundant observations \cite{konidaris2019necessity} and provide sample-efficient behavioral strategies for learning dynamically changing environments.

Several recent approaches attempted to obtain a core state space from full observations. The environment inference-based methods directly estimate the structure of the environment (such as state transition rules) obeying partially observable Markov decision processes (POMDPs) and possibly attain high explanatory power\cite{schrittwieser2020mastering,grimm2021proper}. Other state reduction methods have also been proposed based on state space estimation\cite{ross2007bayes, friston2010free, strens2004efficient, doshi2009infinite}. For instance, factored MDP represents a full observation space as a combination of partial transition rules, assuming that full observations are composed of multiple elements\cite{doya2002multiple, wang2021lifelong}. A low-rank MDP decomposes the transition rule of a fully observed MDP into a low-dimensional matrix\cite{agarwal2020flambe}. It should be noted, however, that these methods reduce the state space based on both rewarded and non-rewarded observations. Therefore, the difficulties associated with redundant observations are only partially resolved and the estimated state space is not necessarily the minimum state space required for an optimal strategy.

Abstract RL\cite{starre2022analysis} uses indices such as behavioral values to reduce the full observation space obeying the Markov decision process (MDP)\cite{bellman1957markovian} to a goal-oriented state space similar to the core state space\cite{abel2018state,abel2019state,allen2021learning}. Deep Q networks are also considered to develop task-specific states representing features necessary to predict rewards efficiently from complex observations\cite{mnih2015human}. These methods rely on black-box function approximation by deep neural networks (DNNs) to facilitate handling complex observations, hence cannot explain why certain behaviors lead to greater rewards\cite{alharin2020reinforcement}.  To improve explainability, MuZero explicitly separates the feature extraction module from the module for estimating the dynamics of the environment and uses the feature extraction module for reward estimation\cite{schrittwieser2020mastering,grimm2021proper}. MuZero has shown state-of-the-art performance in strategy learning using task-specific states.

Nonetheless, DNN learning is inefficient in Lifelong RL, which is a dynamically changing non-stationary environment\cite{chen2018lifelong}. In Lifelong RL, catastrophic forgetting\cite{khetarpal2020towards} forces DNN to relearn the past environments\cite{french1999catastrophic}. Some environment inference-based methods attempt to solve this difficulty using sample-efficient strategy learning, for instance, by switching between multiple modules in response to changes in the environment and adapting to a new environment without forgetting previously learned knowledge\cite{doya2002multiple, wang2021lifelong}. In sum, no effective methods are known to estimate an environment with a priority on reward prediction and adapt to non-stationarity. State-space reduction methods based on environment inference cannot solve the former problem, and DNN-based methods have difficulty in the latter situation. A novel framework of goal-oriented environment inference methods is necessary to attain an optimal estimate of state space in a lifelong RL environment.

To solve the above-mentioned problems, we formulate the redundantly observable MDP (ROMDP) for RL in an environment with many reward-irrelevant observations. Observations for agents can generally be clustered into the core states needed for reward prediction. ROMDP is a form of MDP that conceptually separates observations into core and noise. Therefore, the maximum long-term reward is equally predictable in both cases where the state space consists of all possible observables or only core states. However, the agent can learn an optimal behavioral strategy more effectively if it can identify the core states as these states and their structure (i.e., the transition rule among them) most directly explain the behavioral principles of agents following the optimal strategy. In other words, while the space of all observables is the most redundant state space for reward maximization, the core space gives the minimal state space to maximize the long-term reward.

We show that identifying the core states is crucial for the explainability of the decision-making agent. Furthermore, we show that non-episodic RL in nonstationary ROMDP coupled with dynamically estimating the state space can improve the learning speed of behavioral strategies and follow abrupt changes in the environment. In general, as the redundancy of the state space increases, the number of combinations of states and actions increases, and so does the number of times of trial-and-error for learning an optimal strategy. In contrast, exploration in the core space, which has the fewest combinations of states and actions, will maximize (or at least improve) the learning speed.

To address non-episodic RL in nonstationary ROMDP, we propose the goal-oriented environment inference (GOEI) as a probabilistic model for estimating the core space, and use Action-value Thompson sampling (ATS) for action selection based on the state space structure estimated by the GOEI. We demonstrate that the combination of GOEI and ATS can estimate the state space and simultaneously learn an optimal strategy while retaining the learned knowledge.

\section{Redundantly Observable Markov Decision Process (ROMDP)}
\begin{figure*}[ht]
    \vspace{-0.3cm}
    \centering
    \includegraphics{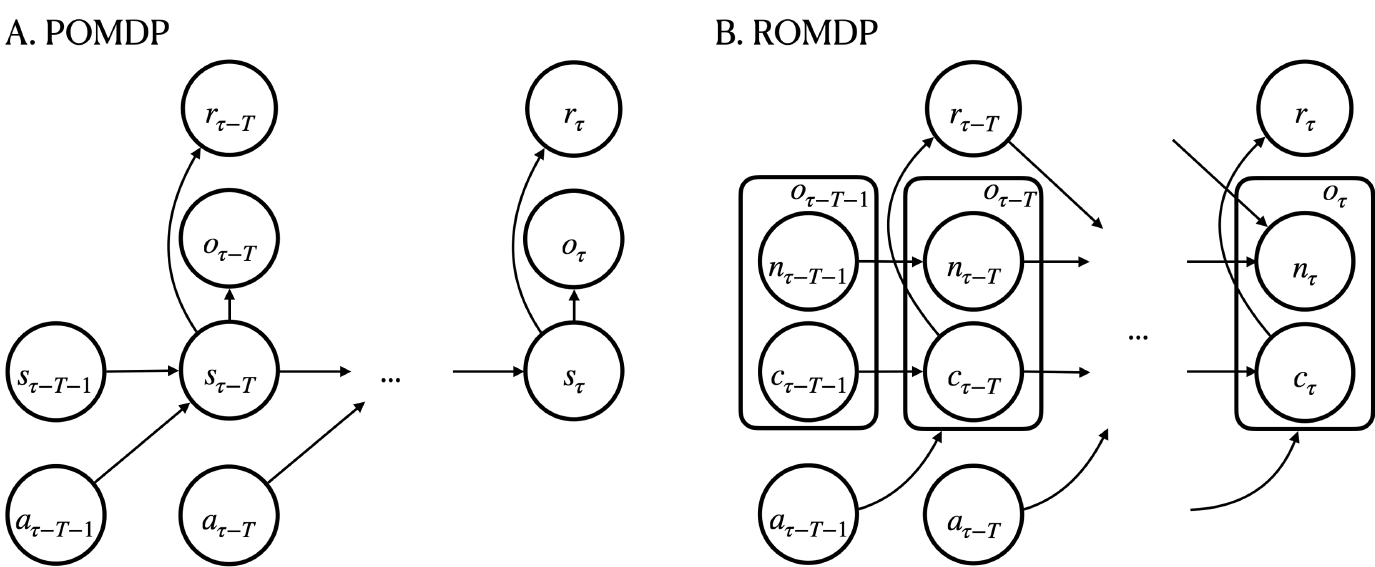}
    \vspace{-0.3cm}
 \caption{Graphical models of POMDP (A) and ROMDP (B). In POMDP, the hidden state $s_t$ determines the action-reward contingency through the action-dependent state transition $p\rbra{s_{t+1}|s_t,a_t}$ and the reward generation rule $p\rbra{r_t|s_t}$. In ROMDP, the observation $o_t$ is a direct product of the core $c_t$ and noise $n_t$, and the core $c_t$ determines the action-reward contingency through the action-dependent core transition $p\rbra{c_{t+1}|c_t,a_t}$ and the reward generation rule $p\rbra{r_t|c_t}$. The noise can vary through the transition rule $p\rbra{n_{t+1}|n_t,a_t,r_t}$ depending on the chosen action and the obtained reward. However, the noise is irrelevant to reward delivery.
 }
    \label{fig:po_romdp}
\end{figure*}
Consider the case that the agent chooses an action $a_t \in A$ for a given observation $o_t \in O$ at time $t$, resulting in reward acquisition $r_{t+1} \in R$.
We will formulate a problem to obtain unknown core states from experiences.
A similar framework is known as the partially observable Markov decision process (POMDP) in which a hidden state is inferred from recent experiences under uncertain observations \cite{ross2007bayes}. 
POMDP assumes that the current observation depends on a hidden state $s_t \in S$ through a conditional probability $p\rbra{o_t|s_t}$ and that $s_t$ determines action-reward contingency through an action-dependent transition rule $p\rbra{s_{t+1}|s_t,a_t}$ and a reward generation rule $p\rbra{r_t|s_t}$ (Fig.~\ref{fig:po_romdp}A). In contrast, ROMDP assumes that the observation $o_t$ is a direct product of ``core'' $c_t \in C$ and ``noise'' $n_t \in N$, and that only the core determines action-reward contingency through an action-dependent transition rule $p\rbra{c_{t+1}|c_t,a_t}$ and a reward generation rule $p\rbra{r_t|c_t}$ (Fig.~\ref{fig:po_romdp}B). Any state irrelevant to reward generation is regarded as noise, which also undergoes transitions $p\rbra{n_{t+1}|n_t,a_t,r_t}$ depending on the chosen action and resultant reward. 
Note that the agent has no a priori knowledge to distinguish between the core and noise. That is, ROMDP can be treated as a special case of POMDP without partial observability, and algorithms supporting POMDPs may find an optimal solution to ROMDP. However, to efficiently attain an optimal behavior, the agent has to cluster the observation into the core through trial and error. 

We further explain the relationship between POMDP and ROMDP. Let $S$ be the minimum state space sufficient to determine action-reward contingency, and we call it goal-oriented state space. In POMDP, the agent has to estimate $S$ from partial observations $O \subseteq S$ whereas in ROMDP the agent estimates the same state space from redundant observations, $S = C \subseteq O$. A combination of POMDP and ROMDP characterizes the most general class of environments, in which the agent is hidden from some observations but is simultaneously exposed to redundant observations. In this case, observation $O$ includes the core observation $C$ and this constitutes the goal-oriented state $S$: $S \cap O = C$. As theoretical frameworks for POMDP first attempts to predict $O$, a more efficient method is required for the effective estimation of $S$ in ROMDP. In this study, we focus on ROMDP and propose a learning method adequate for it. 

Both POMDP and ROMDP satisfy the definition of MDP on the state space of full information. If states are defined with the entire history of a full set of information, i.e., the histories of actions, rewards, and observations, the decision process trivially satisfies the Markov property. Otherway around, the Markov property depends on the definition of state space. POMDP satisfies the Markov property on the essential states extracted from history. ROMDP satisfies the Markov property on the state space defined through full observations. 

\section{Models}

\subsection{Complete Environment Inference (CEI)}

First, we describe the formulation of variational Bayesian inference used in the conventional learning strategy in POMDPs \cite{friston2010free, friston2021sophisticated}, and extend it to a multi-module version for inference in the nonstationary setting. Given the current time $\tau$, consider the series of past states $\tilde{s} = \sbra{s_{\tau - T}, \cdots, s_\tau}$, observations $\tilde{o} = \sbra{o_{\tau - T}, \cdots, o_\tau}$, rewards $\tilde{r} = \sbra{r_{\tau - T}, \cdots, r_\tau}$, actions $\tilde{a} = \sbra{a_{\tau - T -1}, \cdots, a_{\tau-1}}$, parameters of transition rule modules $\tilde{M} = \sbra{M_{\tau-T}, \cdots,M_\tau}$ and parameters of reward rule modules $\tilde{N} = \sbra{N_{\tau-T}, \cdots,N_\tau}$. The generative model for these series is formulated as
\begin{multline}
    p\rbra{\tilde{o}, \tilde{r}, \tilde{s}| \tilde{a}, s_{\tau-T-1}, L, \tilde{M}, \tilde{N}} \\
    = \prod_{t=\tau-T}^{\tau} p\rbra{o_t|s_t, L} p\rbra{s_t|s_{t-1}, a_{t-1}, M_t} p\rbra{r_t|s_t, N_t}, \label{cei}
\end{multline}
where $L, M_t$ and $N_t$ are the parameters of the observation rule $p(o_t|s_t)$, transition rule $p(s_t|s_{t-1}, a_{t-1})$, and reward rule $p(r_t|s_t)$, respectively. Furthermore, $M_t \in \bar{M} = \cbra{M^1, \cdots}$ and $N_t \in \bar{N} = \cbra{N^1, \cdots}$ obey Dirichlet process\cite{blei2006variational} with the hyperparameters ($\alpha_M > 0$ and $\alpha_N >0$) and the base distributions ($B_M$ and $B_N$), respectively.

The objective of Bayesian inference is to obtain the posterior distribution $p\rbra{\tilde{s}, \tilde{M}, \tilde{N}, L, B_M, B_N| \tilde{o}, \tilde{r}, \tilde{a}}$ from the past series $\sbra{\tilde{o}, \tilde{r}, \tilde{a}}$ that the agent has already obtained. However, since it is difficult to analytically obtain the posterior distribution, we calculate an approximate posterior distribution $q\rbra{\tilde{s}, \tilde{M}, \tilde{N}} q(L, B_M, B_N) \simeq p\rbra{\tilde{s},\tilde{M}, \tilde{N}, L, B_M, B_N| \tilde{o}, \tilde{r}, \tilde{a}}$ by using the variational Bayesian method \cite{zhang2018advances,takekawa2009novel}.

If actions, observations, and rewards are discrete, the parameters $L, M_t$ and $N_t$ obey categorical distributions:
\begin{align}
    p\rbra{o_t|s_t, L} &= Cat\rbra{o_t|L\sbra{s_t}}, \\
    p\rbra{s_t|s_{t-1}, a_{t-1}, M_t} &= Cat\rbra{s_t|M_t\sbra{s_{t-1}, a_{t-1}}}, \\
    p\rbra{r_t|s_t, N_t} &= Cat\rbra{r_t|N_t\sbra{s_t}},
\end{align}
respectively.
In this case, the prior distribution of these parameters is the independent distribution with hyperparameters $\Phi^L, \Phi^{M^x}, \Phi^{N^z}, \Phi^{X}$, and $\Phi^{Z}$, respectively: 
\begin{multline}
    p\rbra{L, B_M, B_N} = p\rbra{L}p\rbra{\bar{M}, X} p\rbra{\bar{N}, Z} \\
    = Dir\rbra{L; \Phi^L} \sbra[\big]{\prod_{x}^\infty Dir\rbra{M^x; \Phi^{M^x}} SBP\rbra{X; \Phi^{X}}} \\
    \times \sbra[\big]{\prod_{z}^\infty Dir\rbra{N^z; \Phi^{N^z}} SBP\rbra{Z; \Phi^{Z}}},
\end{multline}
where $X$ and $Z$ are the probabilities for assignment indicator $x \sim Cat\rbra{X}$ and $z \sim Cat\rbra{Z}$ that obey the stick-breaking process (SBP).
The approximate posterior distribution of the parameters is also
given as the product of the independent distributions with the hyperparameters $\Theta^L, \Theta^{M^x}, \Theta^{N^z}, \Theta^{x}$, and $\Theta^{z}$: 
\begin{multline}
    q\rbra{\Omega_{CEI}} := q\rbra{L, \bar{M}, \bar{N}, X, Z} = \\ q\rbra{L} \sbra[\Big]{\prod_x^\infty q\rbra{M^x}} \sbra[\Big]{\prod_z^\infty q\rbra{N^z}} q\rbra{X}q\rbra{Z},
\end{multline}
where
\begin{align}
    q\rbra{L} &= Dir\rbra{L;\Theta^L}, \\
    q\rbra{M^x} &= Dir\rbra{M^x; \Theta^{M^x}}, \\
    q\rbra{N^z} &= Dir\rbra{N^z; \Theta^{N^z}}, \\
    q\rbra{X} &= SBP\rbra{X; \Theta^X}, \\
    q\rbra{Z} &= SBP\rbra{Z; \Theta^Z}.
\end{align}
Then, update rules for the approximate posterior distribution are described as
\begin{multline}
    q\rbra{s_t, M_t, N_t|s_{t-1}} \propto \\
    \exp \cbra{\mathds{E}\sbra{\ln{p\rbra{o_t, r_t, s_t, M_t, N_t|a_{t-1}, s_{t-1}, \Omega_{cei}}}}_{q\rbra{\Omega_{CEI} }}} , \label{update q}
\end{multline}
\begin{equation}
    \Theta^L_{os} \leftarrow \Phi^L_{os} + \sum_{t=\tau-T}^{\tau} q\rbra{s_t=s} \mathds{1}\rbra{o_t=o}, \label{update L}
\end{equation}
\begin{multline}
    \Theta^{M^x}_{s'sa} \leftarrow \Phi^{M^x}_{s'sa} + \\
    \sum_{t=\tau-T}^{\tau} q\rbra{s_t=s', s_{t-1}=s, M_t=M^x}\mathds{1}\rbra{a_{t-1}=a}, \label{update M}
\end{multline}
\begin{gather}
    \Theta^{N^z}_{rs} \leftarrow \Phi^{N^z}_{rs} + \sum_{t=\tau-T}^{\tau} q\rbra{s_t=s, N_t=N^z} \mathds{1}\rbra{r_t=r}, \label{update N}\\
    \Theta^X_x \leftarrow \Phi^X_x + \sum_{t=\tau-T}^{\tau} q\rbra{M_t = M^x}, \label{update F}\\
    \Theta^Z_z \leftarrow \Phi^Z_z + \sum_{t=\tau-T}^{\tau} q\rbra{N_t = N^z}. \label{update G}
\end{gather}
Alternately applying the update rules (\ref{update q}) and (\ref{update L})-(\ref{update G}) leads us to an approximate posterior distribution.

If the above inference for POMDP is applied to ROMDP environment, then the posterior distribution is updated to predict observation from the estimated state. Therefore, when the noise $n_t$ undergoes transitions, the noise can slip into the estimated state space because the redundant observables is predictable. Consequently the estimated state space may not converge to the core space. We call this inference in the ROMDP environment ``Complete Environment Inference (CEI)'' as it tries to predict all observations regardless of whether the estimated state space is the core space.

\subsection{Goal-Oriented Environment Inference (GOEI)}

\begin{figure}[ht]
    \centering
    \includegraphics{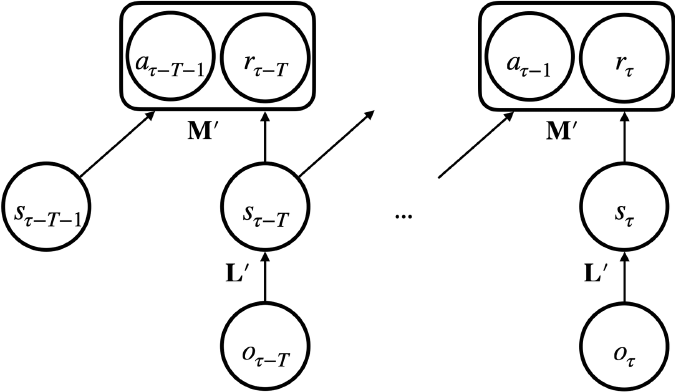}
    \caption{The graphical model of GOEI for ROMDP. The model clusters observations depending on the relationships among states $s_{t-1}$ and $s_t$, action $a_t$, and reward $r_t$. The structure of the model does not match
 that of the ROMDP environment, which is important for the efficient identification of goal-oriented core states.
    }
    \label{fig:goei}
\end{figure}

As CEI indicates, we need an environment inference model that does not predict the observations in ROMDP. Therefore, we introduce ``Goal-Oriented Environment Inference (GOEI)'' that attempts to cluster the redundantly observables into the core as the estimated state $p\rbra{s_t|o_t}$.
However, the clustering rule $p\rbra{s_t|o_t}$ and the transition rule in CEI are incompatible in the Bayesian framework. Therefore, we dare to use a generative model that would not match the environmental structure perse. 

We construct a generative model to describe the relationships between the state transitions that mediate action-reward contingency under the constraint that observations should be only included in the conditions of the conditional probability but not in its outcomes. The resultant generative model has the structure shown in Figure~\ref{fig:goei}:
\begin{multline}
    p\rbra{\tilde{r}, \tilde{a}, \tilde{s}| \tilde{o}, s_{\tau-T-1}
    L', \tilde{M'}} = \\
    \prod_{t=\tau - T}^{\tau}  p\rbra{r_t, a_{t-1}|s_t, s_{t-1}, M'_t}p\rbra{s_t|o_t, L'}. \label{eq:goei}
\end{multline}
The parameters of the clustering rule and the action-reward rule $p\rbra{r_t, a_{t-1}|s_t, s_{t-1}}$ are denoted as $L'$ and ${M'}_t \in \bar{M'}=\sbra{{M'}^1, \cdots}$, respectively, and $M'_t$ obey Dirichlet process with the hyperparameter, $\alpha_{M'} (> 0)$, and the base distribution, $B_{M'}$. This generative model categorizes the observations into clusters that can predict the current action and resultant reward by combining the current and previous clusters $[s_{t-1}, s_t]$. The model $p\rbra{a_{t-1}|s_t, s_{t-1}}$ may look unnatural at first glance, but it only describes how likely the transition $s_{t-1}$ to $s_t$ was caused by the action $a_{t-1}$ when the change of states occurred. The expression does not imply the prediction of the action from the future state. GOEI only learns the environment and plays no direct roles in action selection.

If actions, observations, and rewards are discrete, as in the previous formulation of POMDP, $L'$ and ${M'}_t$ obey categorical distributions. Then, their prior distributions are independent Dirichlet distributions, $p\rbra{L', B_{M'}} = p\rbra{L'; \Phi^{L'}} \sbra{ \prod_y^\infty p\rbra{{M'}^y;\Phi^{{M'}^y}} SBP\rbra{Y; \Phi^Y}}$, and their approximate posterior distribution in the variational Bayesian method is a product of the Dirichlet distribution, $q\rbra{\Omega_{GOEI}}
:= q\rbra{L', \bar{M'}, Y} = q\rbra{L';\Theta^{L'}} \sbra{\prod_y^\infty q\rbra{{M'}^y; \Theta^{{M'}^y}} q\rbra{Y; \Theta^Y}}$, where $y \sim Cat\rbra{Y}$ is the assignment indicator of $M'$. As in CEI, the approximate posterior distribution is updated by alternately applying the following rules (\ref{update qg}) and (\ref{update Lg})-(\ref{update Mg}):
\begin{multline}
    q\rbra{s_t, M'_t|s_{t-1}} \propto \\
    \exp \cbra{\mathds{E}\sbra{\ln{p\rbra{ r_t, a_{t-1}, s_t, M'_t|o_t, s_{t-1}, \Omega_{GOEI}}}}_{q\rbra{\Omega_{GOEI}}}}, \label{update qg}
\end{multline}
\begin{align}
    \Theta^{L'}_{so} &\leftarrow \Phi^{L'}_{so} + \sum_{t=\tau-T}^{\tau} q(s_t=s)\mathds{1}\rbra{o_t=o}, \label{update Lg}
\end{align}
\begin{multline}
    \Theta^{{M'}^y}_{ars's} \leftarrow \Phi^{{M'}^y}_{ars's} + \sum_{t=\tau-T}^{\tau} q\rbra{s_t=s', s_{t-1}=s, M'_t={M'}^y} \\
    \times \mathds{1}\rbra{a_{t-1}=a}\mathds{1}\rbra{r_t=r}, \label{update Mg}
\end{multline}
\begin{equation}
    \Theta^Y_y \leftarrow \Phi^Y_y + \sum_{t=\tau-T}^{\tau} q\rbra{M'_t={M'}^y}. \label{update Hg}
\end{equation}
Importantly, the update of the posterior distribution maximizes the likelihood of action and reward, implying that the model evidence is high even if the observations are unpredictable. Thus, we can achieve GOEI by using a generative model incongruent with the environment.

In order to choose an optimal action according to the optimal bellman equation (Section \ref{QTS}) in each estimated state, the transition rule $p(s_{t+1}|s_t,a_t)$ and reward delivery rule $p(r_t|s_t)$ of the environment are necessary. However, since the GOEI uses the incongruent generative model, it does not directly estimate these rules. Therefore, the transition and reward rules should be inferred independently of the graphical model of GOEI. To overcome this difficulty, we calculate the posterior distributions of the transition rule and reward delivery rule assignment indicator with
\begin{multline}
    q\rbra{M_t|s_t, s_{t-1}} \propto \\
    \exp\cbra{\mathds{E}\sbra{\ln p\rbra{s_t, M_t|s_{t-1}, a_{t-1}, \bar{M}, X}}_{q\rbra{\bar{M}, X}}}, 
\end{multline}
\begin{equation}
    q\rbra{N_t|s_t} \propto \exp\cbra{\mathds{E}\sbra{\ln p\rbra{r_t, N_t|s_t, \bar{N}, Z}_{q\rbra{\bar{N}, Z}}}},
\end{equation}
respectively, and update their rules by substituting the state distribution $q(\tilde{s})$ inferred by iteration of equations (\ref{update qg})-(\ref{update Hg}) into the formula (\ref{update M})-(\ref{update G}).

A virtue of our method is that the agent can initiate inference with a relatively small state space and gradually add novel states based on the necessity. Learning in a limited state space will be more efficient than learning in large state space. The virtue emerges from the fact that the Dirichlet process is applicable to the generative model adopted by our method and the Dirichlet process admits the dynamic growth of the state space. In (\ref{update qg})-(\ref{update Mg}), $s$ and $L'$ obey a categorical distribution and a Dirichlet distribution, respectively. In GOEI, $s$ is to obey a Dirichlet process with hyperparameter $\alpha_{L'} > 0$, $s \sim DP(L', \alpha_{L'})$, allowing an arbitrary number of states during simulations. In CEI, the Dirichlet process cannot be used as the outcomes of $L$ correspond to observations. Thus, the number of states is that of observations.

\subsection{Online execution with forgetting}

To estimate the environment online, we apply the update rules of GOEI at every $T$ step \cite{sato2001online} using the approximate posterior distribution $q(L', B_{M'})$ of updated parameters as the prior distribution $p(L', B_{M'})$ after $T$ steps. The smaller the update range $T$, the faster the trial-and-error estimation of state space. However, reducing the update range $T$ is not trivially easy for two reasons.

Firstly, online updating of the posterior distribution introduces truncation errors. After a truncation of past sequential updates, we cannot re-estimate the truncated past states to improve the current estimation. Since the update of the posterior distribution $q(L', B_{M'})$ depends on the estimated state space $q(s)$, uncertainties in the past state estimation accumulate faster in the posterior distribution for smaller values of $T$. Consequently, the correction of truncation errors in the state space estimation becomes more and more difficult at later steps.

Secondly, every step of updating introduces approximation errors in the approximate posterior distribution. Updating the approximate posterior distribution by a variational Bayesian method assumes the independence of parameters and states. However, this assumption is generally not guaranteed \cite{zhang2018advances} and the approximate posterior distribution accumulates approximation errors.

To reduce the update range, we introduce a forgetting effect in the update rule (\ref{update Lg}). Here, forgetting enhances the effects of observations within the range $T$ in updating the posterior distribution. In other words, forgetting removes accumulated censoring and approximation errors from the posterior distribution by discounting the influence of the prior distribution, which is the memory of the past, on estimating the parameters. The parameter $\eta_t$ dynamically regulates the degree of forgetting according to the correctness of the predicted observation $o_t$ and estimated state $s_t$:
\begin{align}
    \eta_t = (1 - \rho) \left( 1 - q(s_t=s) \right)\mathds{1}(o_t=o), \\
    \Theta^{L'}_{so} \leftarrow (1 - \eta_t) \Theta^{L'}_{so} + q(s_t) \mathds{1}(o_t=o) .
\end{align}
where $\eta_t \in [0, 1 - \rho]$ and $\rho \in [0, 1]$ is the maximum forgetting rate. The parameter  $\Theta_{so}^{L'}$ has initial value $\Phi_{so}^{L'}$ and is sequentially updated from $\tau - T$ to $\tau$. If $\mathds{1}(o_t=o)=1$ and $q(s_t)=0$ when the observation at time $t$ is actually $o$, the maximal forgetting $1 - \eta_t = \rho$ occurs.

\subsection{Action-value Thompson sampling}\label{QTS}

The final goal of the agent is to maximize the amount of acquired reward. If the true state, the transition rule $M^x$, and the reward rule $N^z$ are known, an optimal action pattern in a given state can be determined. In Markov Decision Process, the action pattern that maximizes the action values $\mathcal{Q}(s,a)$ with a discount factor $\gamma<1$ sufficiently close to 1  
\begin{align}
    \mathcal{Q}(s,a) \equiv E\left[\left.\sum_{t=0}^\infty \gamma^t r_{\tau+t+1} \right|s_t=s,a_t=a \right]
\end{align}
is known to maximize the long-term reward \cite{sutton2018reinforcement}. The maximum action values are obtained as the solution of the optimal Bellman equation \cite{watkins1992q}:
\begin{align}
    \vec{\mathcal{Q}}(a_{\tau}) = \left(N^z \vec{r}   
    + \gamma \max_{a_{\tau+1}} \vec{\mathcal{Q}} (a_{\tau+1})\right)^\top M^x_{a_{\tau}} \;. \label{eq:Qvalue}
\end{align}
Here, the$\vec{\mathcal{Q}}(a)$ represents the action values for the respective states $\{s\}$, the $(j,i)$ element of the transition rule $M^x$ is the probability $M_{jia}^x$ of transition from state $s_i$ to $s_j$ by action $a$ and that of the reward rule $N^z$ is the probability $N_{jk}^z$ of obtaining the reward amount $r_k$ in the hidden state $s_j$. The vector $\vec{r}$ represents the reward amounts.

We may solve the Bellman equation by the value iteration method \cite{sutton2018reinforcement}. However, the transition rule, reward rule, their assignment indicator, and true states are unknown to the agent. Therefore, the agent has to estimate them by exploratory behavior, but too much exploration will result in the loss of opportunities to collect rewards. An adequate balance between exploitation and exploration is crucial for learning optimal behavior. We propose to obtain an action policy $a_{\tau}=\arg \max_a \mathds{E}_{q(s_\tau)}\sbra{\mathcal{Q}\rbra{s_\tau,a}}$ by solving the optimal Bellman equation (\ref{eq:Qvalue}) through sampling from the models $M^x$ and $N^z$. Note that both CEI and GOEI use the variational Bayesian inference to estimate the probability distributions $q(B_M)$, $q(B_N)$ and $\vec{q}(s_\tau)$ of the transition rule, reward delivery rule, their assignment indicator, and true states. This enables us to sample the models $M^x$ and $N^z$ from the estimated distributions. 

The proposed action policy will dynamically control exploration and exploitation associated with agent confidence for environment inference. When the accuracy of the estimation is low, the sampled action values from a broad distribution will significantly vary at each time step, leading to exploratory behavior. In contrast, when the accuracy is high, the sampled action values from a narrow distribution will not vary much, resulting in exploitative behavior. The proposed sampling-based action policy extends the Thompson sampling, which is sampling from an estimated reward distribution in the multi-armed bandit problem \cite{agrawal2012analysis}. Therefore, we call the present action policy ``Action-value Thompson Sampling (ATS).''

\section{Numerical experiments}\label{numerical experiment}

\subsection{Core states and noise states}

\begin{figure*}[tb]
    \vspace{-0.cm}
    \centering
    \includegraphics{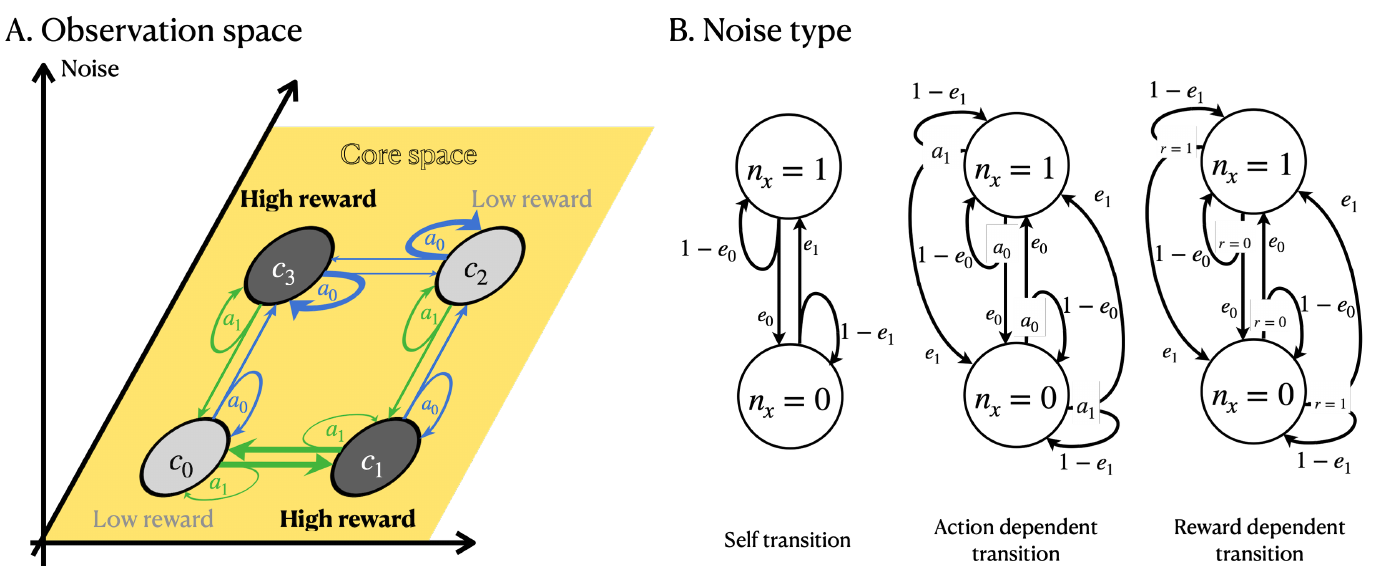}
    \vspace{-0.4cm}
 \caption{Test environments of ROMDP for numerical simulations.
 (A) The space of observations is given as the direct product of core space and noise space orthogonal to it. The core state consists of high-reward (black) and low-reward (grey) states, and the agent can choose between two actions $(a_0,a_1)$. The thickness of the arrows represents the probabilities of the corresponding state transitions. (B) Four types of noise transition rules are considered. Each noise bit is generated through the following transition rules:  $p(n_{t+1}|n_t)$ in the self-transition type; $p(n_{t+1}|n_t, a_t)$ in the action-dependent type; $p(n_{t+1}|n_t, r_t)$ in the reward-dependent type. The parameters $e_0$ and $e_1$ denote the probabilities of the corresponding transitions.
    }\label{fig:problem_setting}
\end{figure*}

\begin{figure*}[ht]
    \centering
    \includegraphics{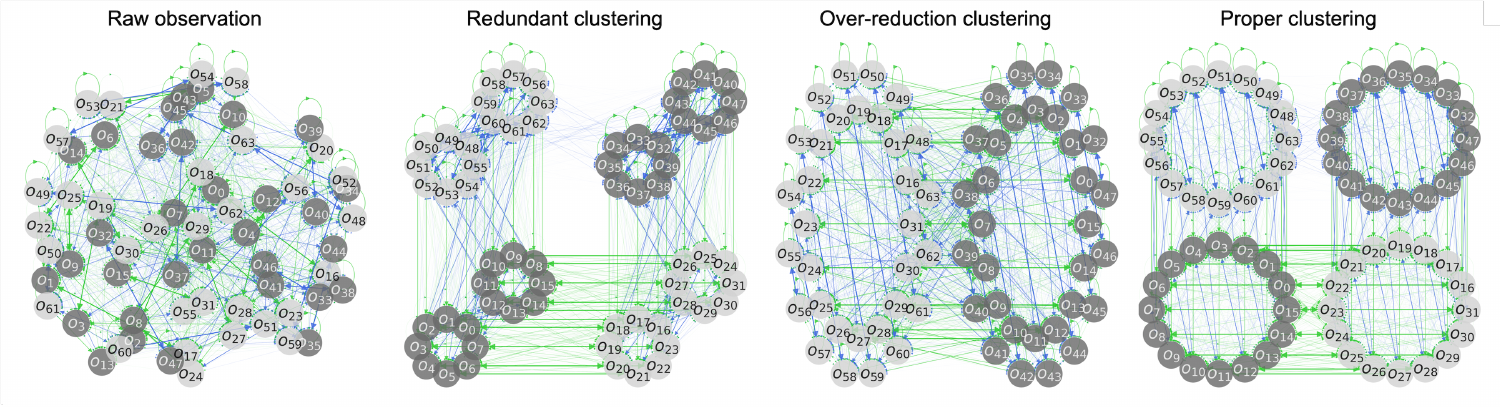}
    \vspace{-0.4cm}
    \caption{
    Clustering results in a task with four core states. Rewards follow the reward rule $N^1$ and state transitions obey the transition rule $M^1$ with 4-bit action-dependent transition noise. Using rewards as a cue, the agent should cluster raw observations obeying a complex transition diagram (leftmost) while preserving the structure of core states. High-reward (dark gray circles) and low-reward (light gray circles) states exist in the transition diagram, and two actions (green, blue) are available for the agent. The rightmost diagrams show correctly clustered four core states, whereas the middle left and right diagrams show redundantly clustered 8 cores and overly clustered two cores, respectively. Observations clustered into a core are arranged in a circle.
    }
    \label{fig:hidden_core}
\end{figure*}

We systematically evaluate the performance of GOEI in the following environments of ROMDP (Figure~\ref{fig:problem_setting}) with nonstationarity. The core space consists of four core states $c_0$, $c_1$, $c_2$, and $c_3$ linked by a transition rule $M^1$, and the agent can visit these states by action selection between $a_0$ and $a_1$. This core space gives a simple example in which the pursuit of immediate rewards is not an optimal policy. To suppress a hasty identification of core states based solely on the magnitudes of single rewards, we deliver a binary reward ($r=1$ or $r=0$) in each core state with a state-dependent reward rule $N_1$. Furthermore, to suppress core state identification solely by reward probabilities, we group the four core states into two core-state pairs, $(c_0, c_1)$ and $(c_2, c_3)$, which have the same combination of reward probabilities. Namely, $p_{N^1,M^1}(r=1|c_1)=p_{N^1,M^1}(r=1|c_3)=0.9$ in high-reward core states and $p_{N^1,M^1}(r=1|c_0)=p_{N^1,M^1}(r=1|c_2)=0.1$ in low-reward core states. The agent can earn a reward most efficiently by selecting the action $a_0$ in the state $c_3$. If this action selection results in a rare transition from $c_3$ to $c_2$, a return trip from $c_2$ to $c_3$ is faster through a round trip via $c_1$ and $c_0$ than directly back to $c_3$. The optimal behavior in this environment is the following pattern: $(c_0, a_0)$, $(c_1, a_1)$, $(c_2, a_1)$, and $(c_3, a_0)$. As the state pairs with the same reward probabilities, i.e., $c_0-c_2$ and $c_1-c_3$, require different actions, the agent cannot learn the optimal policy by discriminating the core states based solely on the reward probabilities.

We introduce $m$-bit noise $(n_1, n_2,\cdots, n_m)$ in the environment, where each noise bit makes transitions between the two states $n_x\in \{0,1\}$. We consider the three types of the transition rules shown in (Figure~\ref{fig:problem_setting}B). The self-transition type makes transitions with the transition rule $p(n_{t+1}|n_t)$. The action-dependent type generates transitions with the rule $p(n_{t+1}|n_t,a_t)$ that depends on the action chosen and the reward-dependent type with the rule $p(n_{t+1}|n_t,r_t)$ that depends on the acquired reward. In the present numerical simulations, the transition probabilities are  $e_0=e_1=0.1$ in the self-transition type, $e_0=0.1$ and $e_1=0.9$ in the action-dependent type, and $e_0=0.1$, $e_1=0.9$ in the reward-dependent type.

Each observation is given as the direct product of core bits and noise bits. Therefore, the number of possible observations is $4\times 2^m$. We represent the observation $o$ with $(4\times 2^m)$-dimensional one-hot vectors to make the estimation of core states from the representations of $o$ difficult. In the present simulations with $m=4$, the task of the agent is to cluster $4\times 2^4=64$ observations into 4 cores. Figure~\ref{fig:hidden_core} schematically explains the different outcomes of clustering raw observations. Proper clustering will give four core states and the transition rules between them. However, the agent may yield more than (redundant clustering) or less than (over-reduction) four clusters in failure cases.

\subsection{Nonstationary test environments} \label{problem_setting}

\begin{figure*}[ht]
 \vspace{-0.3cm}
 \centering
 \includegraphics{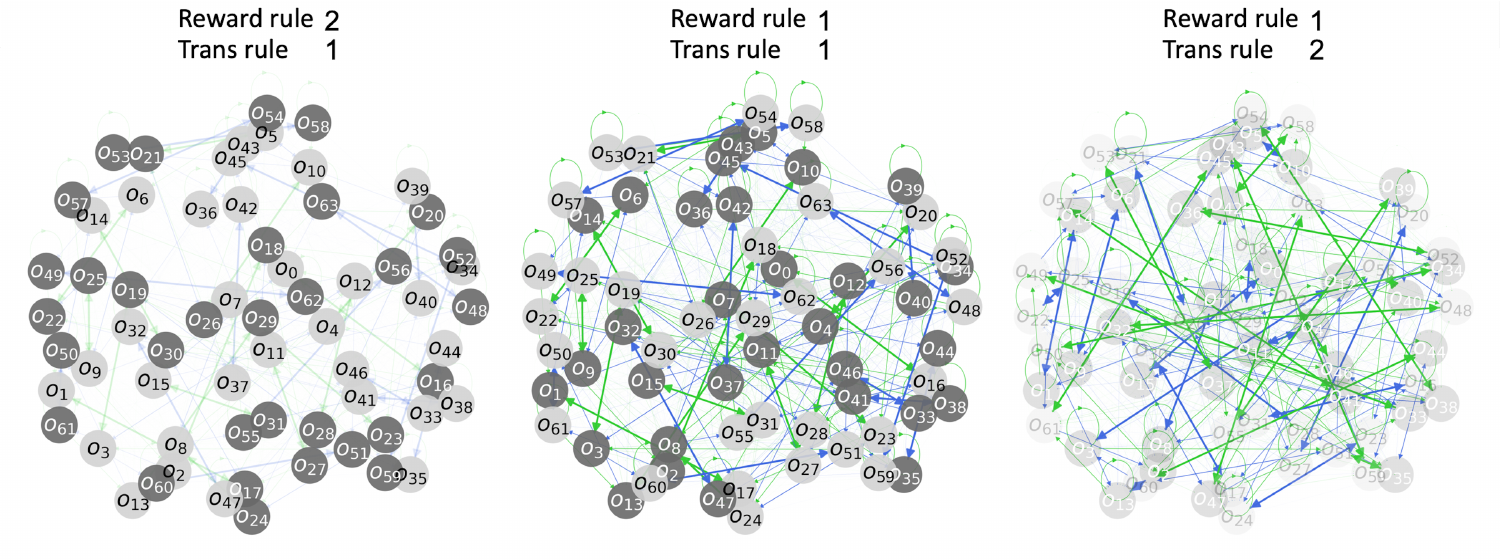}
 \vspace{-0.0cm}
 \caption{
   Observations, including 4-bit action-dependent noise, are indicated by nodes associated with either high rewards (dark gray) or low rewards (light gray). The center panel shows the default transition diagram among the observations obeying the default reward ($N_1$) and transition ($M_1$) rules. The leftmost panel displays the observation diagram when the core reward rule changes to $N_2$, and the rightmost panel shows the transition diagram when the core transition rule changes to $M_2$.
    }
 \label{fig:nonstat_env}
\end{figure*}

We introduce two types of nonstationarity into the base environment defined in the previous subsection. In the nonstationary reward-rule environment, we fix the transition rule at $M^1$ and change the reward rule every 5,000 trials alternately between $N^1$ and $N^2$ (Figure~\ref{fig:nonstat_env}). The reward rule after the change is $p_{N^2,M^1}(r=1|c) = p_{N^1,M^1}(r=0|c)$, and the reward is earned most efficiently by selecting $a_0$ at $c_2$. In contrast, in the nonstationary transition-rule environment, we keep the reward rule $N^1$ unchanged and alternately switch the transition rule between $M^1$ and $M^2$ every 5000 trials. The transition rule after the change is 
\begin{multline}
p_{N^1,M^2}(c|c, \cbra{a_0, a_1}) = \\
\begin{cases}
1 - p_{N^1,M^1}(c|c, \cbra{a_1, a_0}) & p_{N^1,M^1}(c|c, \cbra{a_1, a_0}) \neq 0 \\
0 & p_{N^1,M^1}(c|c, \cbra{a_1, a_0}) = 0
\end{cases}
,
\end{multline}
and selecting $a_0$ at $c_1$ yields the optimal strategy to earn rewards. In both environments, the common optimal behavioral pattern is given as ($(c_0, a_1)$, $(c_1, a_0)$, $(c_2, a_0)$, and $(c_3, a_1)$).

\subsection{Evaluation of performance}
The goal of ROMDP is to estimate the minimum state space that allows an optimal behavior, that is, to reduce as much information as possible from the observation space while keeping all the information about the core space. To evaluate the degree of this reduction, we define information loss and information reduction using the conditional entropy $H(\cdot|s)\ge 0$,
\begin{align}
\mbox{Information loss: }    &H(c|s) = - \sum_{c \in C} \sum_{s \in S} p(c, s) \ln{p(c|s)},\\
\mbox{Information reduction: } &H(o|s) = - \sum_{o \in O} \sum_{s \in S} p(o, s) \ln{p(o|s)},
\end{align} 
where $s$ is the internal state estimated by the agent, and $c$ is the core state, which exists in the environment but is unknown to the agent. The information loss $H(c|s)=0$ when the estimated state $s$ has all the information in the core space, and it increases as the core information is lost. When the estimated state represents all observations, the information reduction $H(o|s) = 0$ and increases as information is reduced about some redundant observations. The core space can be represented with fewer states when $H(o|s)$ can be further maximized while keeping $H(c|s)=0$. The maximum value of $H(o|s)$ depends on the distribution of observations. In our test environments shown in Fig.~\ref{fig:problem_setting}, the dimension of noise space is $m=4$ and the theoretical upper bound is $H(o|s)\fallingdotseq 4.15$.

\subsection{Results}\label{result}

Combining state space estimation with GOEI and action policy with ATS, we addressed the relationship between the degree of state space reduction and the speed of strategy learning. We compared the results with those obtained by the combination of CEI and ATS. The update frequency and maximum forgetting rate were adjusted to most efficiently estimate the state space that can maximize the long-term reward: $T=500$, $\rho=0.95$. The discount rate was set as $\gamma=0.95$. We note that the agent had no prior knowledge about the structure of observations in the simulations below.

\begin{figure*}[ht]
\vspace{-0.3cm}
 \centering
 \includegraphics{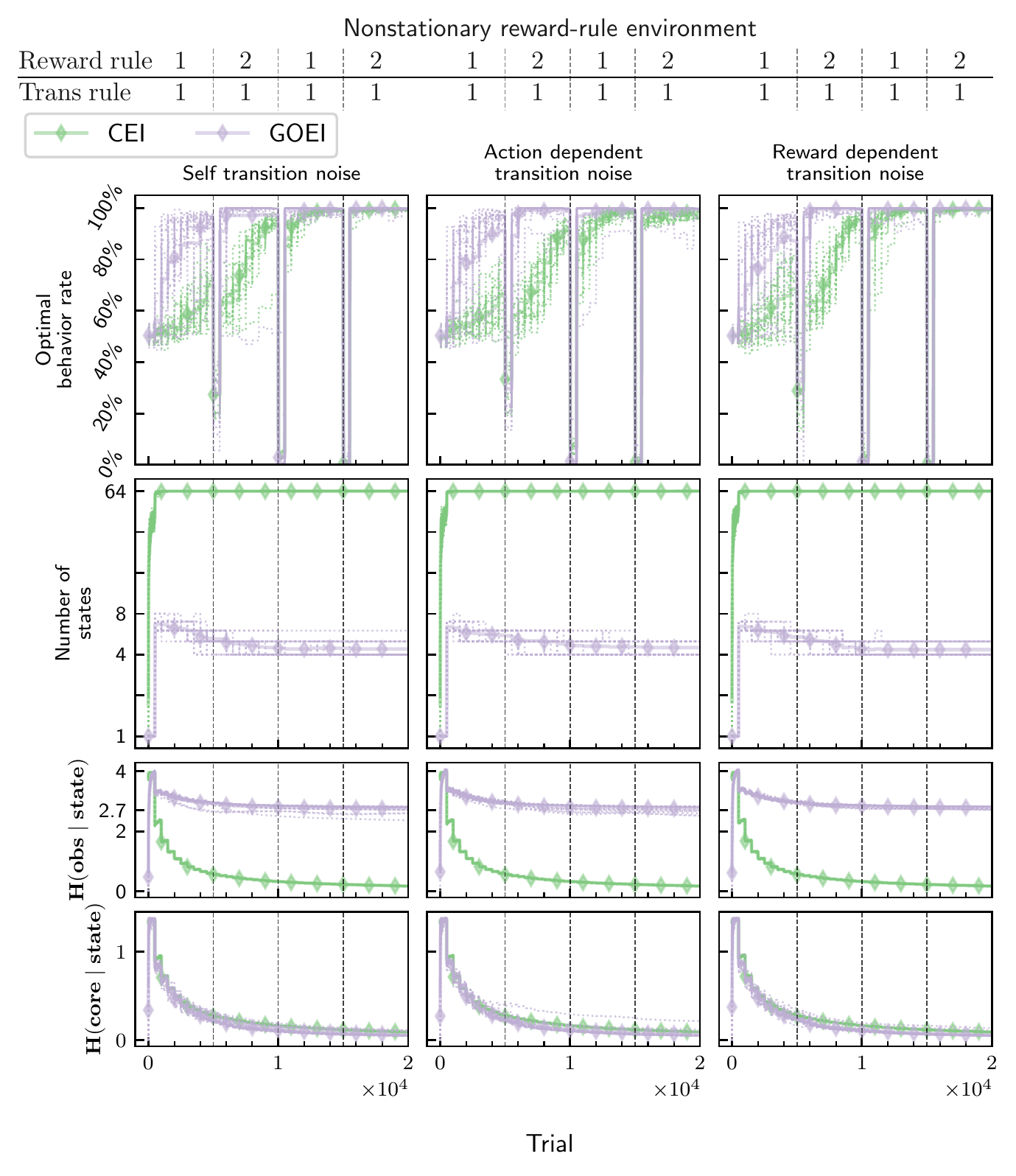}
 \vspace{-0.8cm}
  \caption{Comparisons between GOEI and CEI in nonstationary reward-rule environments of ROMDP. Two reward rules were alternately applied every 5000 trials, as illustrated in the top row. Simulations were performed for three noise types. The numbers of all possible observations and cores were $64$ and $4$, respectively. The optimal behavior rate refers to the number ratio of optimal actions to all actions taken by the agent during $T=500$ trials. The state space is most redundant when the number of states is $64$. In each panel, dashed lines show the results for individual samples, and solid lines marked with diamonds show the average over 20 samples.}
  \label{fig:reward_nonstat_result}
\end{figure*}

\begin{figure*}[ht]
\vspace{-0.3cm}
 \centering
 \includegraphics{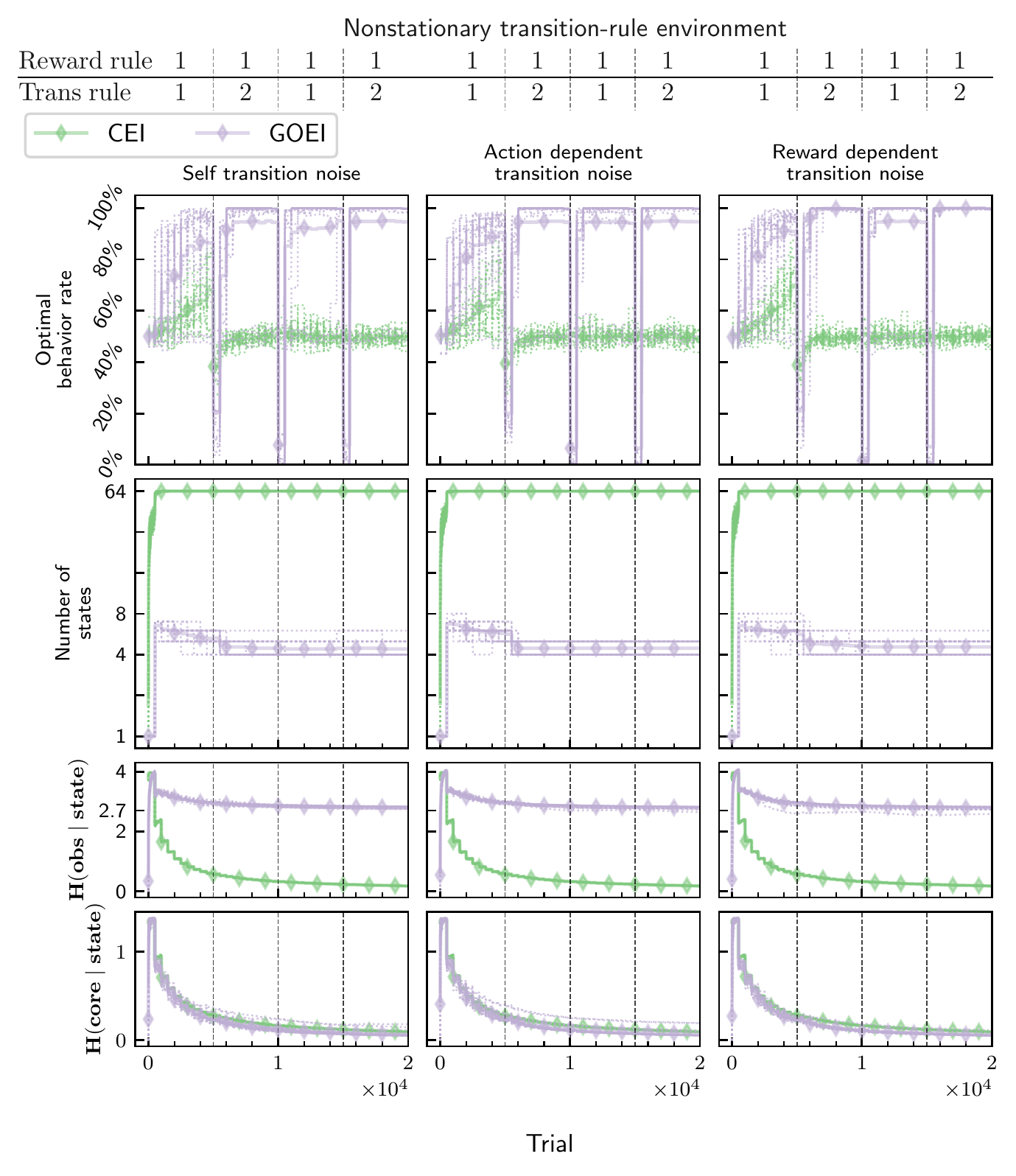}
 \vspace{-0.8cm}
  \caption{Comparisons between GOEI and CEI in nonstationary transition-rule environments of ROMDP. Two state-transition rules were alternately applied every 5000 trials, as indicated in the top row. Other simulation settings were the same as those used in Fig.~\ref{fig:reward_nonstat_result}.}
  \label{fig:trans_nonstat_result}
\end{figure*}

In both non-stationary reward-rule (Fig.~\ref{fig:reward_nonstat_result}, the second row) and transition-rule  (Fig.~\ref{fig:trans_nonstat_result}, the second row) environments, GOEI could always estimate a state space sufficiently close to the core space for all noise types.  The degree of reduction in the estimated state space was given as $H(o|s)\fallingdotseq 2.7$ (Fig. ~\ref{fig:reward_nonstat_result} and \ref{fig:trans_nonstat_result}, the third row) and $H(c|s)=0$ (the fourth row). These results indicate that the state space was appropriately reduced without losing information about the core states. In contrast, CEI could not correctly estimate the state space for all noise types: the estimated number of states was not $4$ but $64$, which was the total number of observations. Further, the prediction accuracy was improved only after the estimated state space was enlarged (Fig.~\ref{fig:reward_nonstat_result} and \ref{fig:trans_nonstat_result}, the second and third rows). Thus, GOEI outperformed CEI in the estimation of core states.

In GOEI, the appropriate reduction of the state space also increased online learning speed and the performance of inferring nonstationarity of transition rules and earning rewards. In the top rows of Fig.~\ref{fig:reward_nonstat_result} and \ref{fig:trans_nonstat_result}, we compared the ratio of optimal actions to the total number of actions between GOEI and CEI. In the first 5000 trials of GOEI, the optimal action ratio steadily approached $100^\%$ as the number of states was reduced. GOEI was able to follow the rule change in both nonstationary reward-rule and transition-rule environments without forgetting the learned knowledge of the environmental structures. Similarly, CEI managed to follow the environmental changes in the reward rule (Fig.~\ref{fig:reward_nonstat_result}). However, CEI failed to follow the changes in the transition rule (Fig. \ref{fig:trans_nonstat_result}). In non-stationary uncertain environments, the agent must judge whether the observed inconsistency between the model's estimation and the obtained data indicates actual changes in the environmental rules or merely uncertainty in observations. For the updating width of $T=500$, each estimated state in GOEI on average gives $500/4= 125$ clues for this judgment, whereas each state in CEI can only provide $500/64 \fallingdotseq 7.8$ clues. The small number of clues makes it difficult for CEI to distinguish between environmental changes and uncertain observations. Therefore, compact state space is crucial for robustly estimating dynamically changing environments.

\section{Discussion and conclusion}
Reinforcement learning is difficult and time-consuming in real-world environments partly because the state space essential to achieving a behavioral goal is unknown to the agent. Learning an optimal policy is easier if the agent can infer this state space accurately from limited experience. This motivated us to extend reinforcement learning to a novel class of reinforcement learning tasks, that is, ROMDP. Unlike the well-known POMDP, ROMDP yields redundant observations which are irrelevant to rewards. Developing efficient methods to solve ROMDP is a crucial step toward solving the general class of reinforcement learning tasks. 

Deep learning can learn an optimal behavioral strategy even in a complex environment by predicting the outcomes of actions through a black box of state space. Such a black box is learnable if big data are available. However, this is not always the case, and the black box cannot tell why the learned strategy works well. When the amount of experiences is limited, such as in a novel environment, the explainability of the past learning effects is important. The explainability is almost straightforward if the agent explicitly learns the minimum state space necessary for achieving a goal.

We proposed GOEI to extract this minimum state space. A characteristic of GOEI is that it uses a generative model with a structure different from that of the generative process of observations in the environment of ROMDP. This incongruent generative model enables memory-efficient goal-oriented inference based on the Dirichlet process: the process allows a substantially small initial state space for the environment inference.

The free energy principle\cite{friston2010free} provides a framework
for environment inference, which is equivalent to variational Bayesian inference, and has been successful in POMDPs. This method does not rely on a black box, well estimates the hidden states essential for learning, is highly explainable and can learn an optimal behavior from limited experience as far as the number of redundant observations is small. However, the previous applications of this method were formulated to predict a complete set of observations and did not aim to reduce the number of necessary states for optimal behavior.

Our results show that modeling the causal relationships between observations in the essential minimum state space (i.e., core space) is effective for learning from limited experience. Actually, cases are known in which a reward reduces the state space of human subjects in a multi-armed bandit problem involving redundant observations \cite{cortese2021value}. We may interpret our method for estimating the minimum state space as the free-energy principle for perception and action based on an incongruent generative model.

Another interesting difference exists between the free energy principle and our method. The former often adopts active inference to generate
actions in an inference-oriented manner. In contrast, our method determines the action policy by solving the optimal Bellman equation in a goal-oriented manner. Further, to regulate the exploration-exploitation balance, our method adopts ATS (action-value Thompson sampling) for sampling states and actions from the estimated environment model. Importantly, the combined use of GOEI and ATS enables an online estimation of the environment. Thus, the free energy principle and our method are different in their action generations. However, they both balance exploration and exploitation, so a theoretical connection may exist between ATS and active inference.

\section{Acknowledgments}
This work was supported by KAKENHI nos. JP18H05213, JP19H04994 and JP18H05524 from JSPS.

\appendix
\section{Theoretical result}

\subsection{Variational bayes}\label{VB}

To illustrate the variational Bayesian process commonly used by CEI and GOEI,
we denote the targets of the model, the given conditions,
and the parameters as $D$, $\bar{D}$, and $\Omega$, respectively.
Under this notation, the generative model is
$\cond*{p}{\tilde{s}, D}{s_{\tau-T-1}, \Omega, \bar{D}}$,
and the prior distributions are 
$\prob*{p}{s_{\tau-T-1}}$, and $\prob*{p}{\Omega}$.
Variational Bayes (VB) assumes that
the posterior distribution of the hidden state $\tilde{s}$ and
the parameters $\Omega$ are independent.
The free energy (FE) is then defined
for the environment estimation model
$\cond*{p}{\tilde{s}, \Omega, D}{s_{\tau-T-1}, \bar{D}}
= \cond*{p}{\tilde{s}, D}{s_{\tau-T-1}, \Omega, \bar{D}}
\, \prob*{p}{\Omega}$
and the approximate posterior distribution
$\prob*{q}{\tilde{s}}\,\prob*{q}{\Omega}$:
\begin{multline}\label{eq:FE}
    \mathrm{FE}
    = \sum_{}
    {\prob*{q}{\tilde{s}}\,\prob*{q}{\Omega}}\,
    \ln\frac{\prob*{q}{\tilde{s}}\,q(\Omega)}{\cond*{p}{\tilde{s}, \Omega, D}{s_{\tau-T-1}, \bar{D}}}
    \\
    = \KL[\Big]{\prob*{q}{\tilde{s}}\,q(\Omega)}
    {\cond*{p}{\tilde{s},\Omega}{D, s_{\tau-T-1}, \bar{D}}} \\
    - \ln{\cond*{p}{D}{s_{\tau-T-1}, \bar{D}}},
\end{multline}
where $\KL{\cdot}{\cdot}$ is the Kalback-Leibler (KL) divergence.

Minimizing FE is equivalent to minimizing the KL-divergence of the approximate posterior distribution and the true posterior distribution.
By the variational theorem, we can define E-step (Equation~(\ref{estep})) that minimizes FE given $\prob*{q}{\tilde{s}}$ and M-step (Equation~(\ref{mstep})) that minimizes FE given $\prob*{q}{\tilde{s}}$.
In variational Bayes, E-step and M-step are iterated alternately to decrease FE continuously:
\begin{align}
    \prob*{q^{(i)}}{\tilde{s}}
    &\propto \exp\sbra*{\E*{q^{(i)}(\Omega)}{\ln \cond*{p}{\tilde{s}, D}{\Omega, s_{\tau-T-1}, \bar{D}}}}
    , \label{estep}
    \\
    \prob*{q^{(i+1)}}{\Omega}
    &\propto \exp\sbra*{\E*{\prob*{q^{(i)}}{\tilde{s}}}{\ln p\rset*{\tilde{s}, D}{\Omega, s_{\tau-T-1}, \bar{D}}}}
    \; p(\Omega). \label{mstep}
\end{align}

In this paper, the posterior distribution obtained in the previous learning was used as the initial condition in VB:
$\prob*{q_{\tau}^{(0)}}{\Omega} = \prob*{p_{\tau}}{\Omega} = \prob*{q_{\tau-T}^{(\infty)}}{\Omega}$.

\subsection{SBP and VB} \label{SBP}
Define $\mathcal{F}$ that follows the Dirichlet process of the base distribution $B$ using SBP as follows:
\begin{gather}
    p\rbra{B} = \prod_\mathcal{X}^\infty p\rbra{F^\mathcal{X};\Phi^{F^\mathcal{X}}} SBP\rbra{\mathcal{X};\Phi^\mathcal{X}}, \\
    p\rbra{v_\mathcal{X}} = \cond*{\mathrm{Beta}}{v_{\mathcal{X}}}{1 + \Phi^{\mathcal{X}}, \alpha_\mathcal{F} + \sum_{i=\mathcal{X}+1}^\infty \Phi^i},\label{sbp1} \\
    SBP\rbra{\mathcal{X};\Phi^\mathcal{X}} = p\rbra{v_\mathcal{X}} \prod_{i=1}^{\mathcal{X}-1} \rbra{1 - p\rbra{v_i}}, \label{ld}
\end{gather}
where $\alpha_\mathcal{F} > 0$ is SBP parameter.
As a countermeasure for the problem that exchangeability does not generally hold in SBP, $\mathcal{X}$ is sorted in descending order of the prior distribution $\Phi^{\mathcal{X}}$ and then Equations (\ref{sbp1}) and (\ref{ld}) is applied.

In m-step, the hyperparameter $\Theta^\mathcal{F}_\mathcal{X}$ of the posterior distribution is updated using the following formula:

\begin{multline}
    \sum_{t=\tau-T}^\tau \prob*{q}{F_t} \ln SBP\rbra{\mathcal{X};\Phi^\mathcal{X}} = \\
    \sum_{t=\tau-T}^\tau \prob*{q}{F_t=F^\mathcal{X}} \ln p\rbra{v_\mathcal{X}} \\
    + \sum_{\mathcal{X}'=\mathcal{X}+1}^\infty \sum_{t=\tau-T}^\tau \prob*{q}{F_t=F^{\mathcal{X}'}} \ln \rbra*{1-p\rbra{v_{\mathcal{X}'}}}.
\end{multline}
In addition, e-step calculates the expected value of the hidden variable using the following equation:
\begin{multline}
\E[\Big]{q\rbra{B}}{\ln SBP\rbra{\mathcal{X}}}
    = \sbra*{\psi\rbra{\Theta^V_\mathcal{X}}-\psi\rbra{\Theta^V_\mathcal{X} + \bar{\Theta}^V_\mathcal{X}}} \\
    \times \sum_{k=1}^{\mathcal{X}-1}\sbra*{\psi\rbra{\bar{\Theta}^V_k} - \psi\rbra{\Theta^V_k + \bar{\Theta}^V_k}},\\
\end{multline}
where $1 + \Theta^\mathcal{X} = \Theta^V_\mathcal{X}$ and 
$\alpha + \sum_{i=\mathcal{X}+1}^\infty \Theta^i = \bar{\Theta}^V_\mathcal{X}$,
for simplicity on notation.

\subsection{Complete environment inference and VB}
Here, we derive an updated formula for the approximate posterior distribution of CEI with VB.
CEI predicts reward $r$ and observation $o$, but not action $a$.
In summary, the definitions in section\ref{VB} are $D=\cbra*{\tilde{r}, \tilde{o}}$, $\bar{D}=\cbra*{\tilde{a}}$ and $\Omega = \cbra*{L, B_M, B_N}$.

For M-step, Equation(\ref{cei}) is substituted into Equation~(\ref{mstep}):
\begin{multline}
    \prob*{q}{L, B_M, B_N} = \\
    \exp\sbra[\Bigg]{ \sum_{t=\tau-T}^\tau \prob*{q}{s_t, M_t, N_t} \cbra[\Big]{
        \ln L_{o_t s_t}
        + \ln M_{s_t s_{t-1} a_{t-1}}^{x_t} \\
        + \ln N_{r_t s_t}^{z_t} 
        + \ln SBP\rbra{x_t}
        + \ln SBP\rbra{z_t}
    }}
    \, \prob*{p}{L, B_M, B_N}.
\end{multline}
The update equations for the hyperparameters of each Dirichlet distribution are as Equations~(\ref{update L})-(\ref{update G}).

Substituting Equation~(\ref{cei}) into Equation (\ref{estep}),
E-step can be easily decomposed into $\prob*{q}{\tilde{s}, \tilde{M}, \tilde{N}} = \prod_{t=\tau-T}^\tau \cond*{q}{s_t, M_t, N_t}{s_{t-1}}$:
\begin{multline}
    \cond*{q}{s_t, M_t, N_t}{s_{t-1}} \propto 
    \exp\cbra[\Bigg]{
        \E[\big]{\prob*{q}{L}}{\ln L_{o_t s_t}} \\
        +
        \E[\big]{\prob*{q}{M^{x}}}{\ln M_{s_t s_{t-1} a_{t-1}}^{x_t}}
        +
        \E[\big]{\prob*{q}{N^{z}}}{\ln N_{r_t s_t}^{z_t}} \\
        +
        \E[\big]{\prob*{q}{B_M}}{\ln SBP\rbra{x_t}}
        +
        \E[\big]{\prob*{q}{B_N}}{\ln SBP\rbra{z_t}}
    }.
\end{multline}
Therefore, the posterior probability of the state at time $\tau-T$
can be first obtained using the prior distribution $\prob*{p}{s_{\tau-T-1}}$
by $\prob*{q}{s_{\tau-T}, M_{\tau-T}, N_{\tau-T}} = \sum_{s_{\tau-T-1}}\cond*{q}{s_{\tau-T}, M_{\tau-T}, N_{\tau-T}}{s_{\tau-T-1}}\,\prob{p}{s_{\tau-T-1}}$.
Then, the simultaneous probability
$\prob*{q}{s_{t-1}, s_t, M_t, N_t} = \cond*{q}{s_t, M_t, N_t}{s_{t-1}}\,\prob*{q}{s_{t-1}}$
and the marginalized probability
$\prob*{q}{s_t} = \sum_{s_{t-1}, M_{t}, N_t} \prob*{q}{s_{t-1}, s_t, M_t, N_t}$
at each time $t$ can be obtained in order from $\tau-T+1$ to $\tau$.

The expected value for the logarithm of a variable following a Dirichlet distribution can be calculated by the digamma function $\psi(\cdot)$:
\begin{align}
    \E[\big]{q(L)}{\ln L_{os}}
    &= {\psi\rbra*{\Theta_{o s}^L} - \psi\rbra*{\sum_{o} \Theta_{o s}^L}}
    \\
    \E[\big]{q(M^x)}{\ln M_{s's a}^{x}}
    &= {\psi\rbra*{\Theta_{s' s a}^{M^x}} - \psi\rbra*{\sum_{s'} \Theta_{s' s a}^{M^x}}}, \\
    \E[\big]{q(N^z)}{\ln N_{rs}^{z}}
    &= {\psi\rbra*{\Theta_{r s}^{N^z}} - \psi\rbra*{\sum_{r} \Theta_{r s}^{N^z}}}, \\
    \E[\big]{\prob*{q}{B_M}}{\ln SBP\rbra{x_t}}
    &=
    \sbra*{\psi(\Theta^V_{x})-\psi(\Theta^V_{x} + \bar{\Theta}_{x}^V)} \nonumber \\
    &\times \sum_{k=1}^{x-1}\sbra*{\psi (\bar{\Theta}_{k}^V) - \psi(\Theta^V_{k} + \bar{\Theta}^V_{k})},\\
    \E[\big]{\prob*{q}{B_N}}{\ln SBP\rbra{z}}
    &=
    \sbra*{\psi(\Theta^V_{z})-\psi(\Theta^V_{z} + \bar{\Theta}_{z}^V)} \nonumber \\
    &\times \sum_{k=1}^{z-1}\sbra*{\psi (\bar{\Theta}_{k}^V) - \psi(\Theta^V_{k} + \bar{\Theta}^V_{k})}.
\end{align}

\subsection{Goal-oriented environment inference and VB}

We derive an updated formula for the approximate posterior distribution by VB for GOEI.
GOEI predicts rewards $r$ and action $a$ and does not predict the observation $o$.
In summary, the definitions in subsection \ref{VB} are
$D=\cbra{\tilde{a}, \tilde{r}}$,
$\bar{D}=\cbra{\tilde{o}}$,
and $\Omega=\cbra{L', B_{M'}}$.

Substituting Equation~(\ref{eq:goei}) into Equation~(\ref{mstep}),
we obtain the M-step of GOEI as:
\begin{multline}
    \prob*{q}{L', B_{M'}} = \\ \exp\sbra[\Bigg]{ \sum_{t=\tau-T}^\tau \prob*{q}{s_t, M'_t} \cbra[\Big]{
        \ln {L'}_{s_t}^{o_t}
        +
        \ln {M'}_{a_{t-1}r_t s_t s_{t-1}}^{y_t} \\
        + \ln SBP\rbra{y_t}
    }}
    \; \prob*{p}{\hat{L'}, B_{M'}}.
\end{multline}
The update formula for the hyperparameters are as Equations~(\ref{update Lg})-(\ref{update Hg}).

The E-step of GOEI can be obtained by substituting Equation~(\ref{eq:goei}) into Equation~(\ref{estep}) as with CEI.
As in CEI case, E-step of GOEI can be easily decomposed into $\prob*{q}{\tilde{s}, \tilde{M'}} = \prod_{\tau-T}^\tau \cond*{q}{s_t, M'_t}{s_{t-1}}$:
\begin{multline}
    \cond*{q}{s_t, M'_t}{s_{t-1}}
    \propto \\ 
    \exp\sbra[\Bigg]{\E[\Big]{q\rbra{L'}}{\ln {L'}_{s_t}^{o_t}}
    + \E[\Big]{q\rbra*{M'}}{\ln {M'}_{a_{t-1}r_t s_t s_{t-1}}^{y_t}} \\
    + \E[\Big]{q\rbra*{B_{M'}}}{\ln SBP\rbra{y_t}}
    },
\end{multline}
resulting in $\prob*{q}{s_{t-1}, s_t, M'_t}$ and $\prob*{q}{s_t}$.

The expected value required by the E-step is obtained as follows:
\begin{align}
    \E[\Big]{q\rbra{L'}}{\ln {L'}_{s}^{o}}
    &=
    \sbra*{\psi(V^{L'}_{s o})-\psi(V^{L'}_{s o} + \bar{V}^{L'}_{s o})} \nonumber \\
    &\times \sum_{k=1}^{s-1}\sbra*{\psi (\bar{V}^{L'}_{k o}) - \psi(V^{L'}_{k o} + \bar{V}^{L'}_{k o})}
    ,
    \\
    \E[\Big]{\prob*{q}{M'}}{\ln {M'}^y_{a r s' s}}
    &= 
    \prob*{\psi}{\Theta_{a r s' s}^{{M'}^y}} - \prob*{\psi}{\sum_{a, r} \Theta_{a r s' s}^{{M'}^y}}, \\
    \E[\big]{\prob*{q}{B_{M'}}}{\ln SBP\rbra{y}}
    &=
    \sbra*{\psi(\Theta^V_{y})-\psi(\Theta^V_{y} + \bar{\Theta}_{y}^V)} \nonumber \\
    &\times \sum_{k=1}^{y-1}\sbra*{\psi (\bar{\Theta}_{k}^V) - \psi(\Theta^V_{k} + \bar{\Theta}^V_{k})}
\end{align}

By introducing the Dirichlet Process, the number of states $|s|$ becomes variable. The number of states $|s|$ becomes variable and The Dirichlet Distribution is assumed and the number of states is $\E[\Big]{q\rbra{L'}}{\ln {L'}_{s}^{o}}$ is different from that assuming Dirichlet Distribution.

\bibliographystyle{unsrt}
\bibliography{reference.bib}

\begin{thebibliography}{10}

\bibitem{mnih2013playing}
Volodymyr Mnih, Koray Kavukcuoglu, David Silver, Alex Graves, Ioannis
  Antonoglou, Daan Wierstra, and Martin Riedmiller.
\newblock Playing atari with deep reinforcement learning.
\newblock {\em arXiv preprint arXiv:1312.5602}, 2013.

\bibitem{DBLP:journals/corr/HausknechtS15}
Matthew~J. Hausknecht and Peter Stone.
\newblock Deep recurrent {Q}-learning for partially observable {MDP}s.
\newblock {\em CoRR}, abs/1507.06527, 2015.

\bibitem{konidaris2019necessity}
George Konidaris.
\newblock On the necessity of abstraction.
\newblock {\em Current Opinion in Behavioral Sciences}, 29:1--7, 2019.

\bibitem{schrittwieser2020mastering}
Julian Schrittwieser, Ioannis Antonoglou, Thomas Hubert, Karen Simonyan,
  Laurent Sifre, Simon Schmitt, Arthur Guez, Edward Lockhart, Demis Hassabis,
  Thore Graepel, et~al.
\newblock Mastering atari, go, chess and shogi by planning with a learned
  model.
\newblock {\em Nature}, 588(7839):604--609, 2020.

\bibitem{grimm2021proper}
Christopher Grimm, Andr{\'e} Barreto, Greg Farquhar, David Silver, and Satinder
  Singh.
\newblock Proper value equivalence.
\newblock {\em Advances in Neural Information Processing Systems},
  34:7773--7786, 2021.

\bibitem{ross2007bayes}
Stephane Ross, Brahim Chaib-draa, and Joelle Pineau.
\newblock Bayes-adaptive {POMDP}s.
\newblock In {\em NIPS}, pages 1225--1232, 2007.

\bibitem{friston2010free}
Karl Friston.
\newblock The free-energy principle: a unified brain theory?
\newblock {\em Nature Reviews Neuroscience}, 11(2):127--138, 2010.

\bibitem{strens2004efficient}
Malcolm Strens.
\newblock Efficient hierarchical {MCMC} for policy search.
\newblock In {\em ICML}, page~97, 2004.

\bibitem{doshi2009infinite}
Finale Doshi-Velez.
\newblock The infinite partially observable markov decision process.
\newblock {\em Advances in neural information processing systems}, 22, 2009.

\bibitem{doya2002multiple}
Kenji Doya, Kazuyuki Samejima, Ken-ichi Katagiri, and Mitsuo Kawato.
\newblock Multiple model-based reinforcement learning.
\newblock {\em Neural computation}, 14(6):1347--1369, 2002.

\bibitem{wang2021lifelong}
Zhi Wang, Chunlin Chen, and Daoyi Dong.
\newblock Lifelong incremental reinforcement learning with online bayesian
  inference.
\newblock {\em IEEE Transactions on Neural Networks and Learning Systems},
  2021.

\bibitem{agarwal2020flambe}
Alekh Agarwal, Sham Kakade, Akshay Krishnamurthy, and Wen Sun.
\newblock Flambe: Structural complexity and representation learning of low rank
  mdps.
\newblock {\em Advances in neural information processing systems},
  33:20095--20107, 2020.

\bibitem{starre2022analysis}
Rolf~AN Starre, Marco Loog, and Frans~A Oliehoek.
\newblock An analysis of abstracted model-based reinforcement learning.
\newblock {\em arXiv preprint arXiv:2208.14407}, 2022.

\bibitem{bellman1957markovian}
Richard Bellman.
\newblock A markovian decision process.
\newblock {\em Journal of mathematics and mechanics}, pages 679--684, 1957.

\bibitem{abel2018state}
David Abel, Dilip Arumugam, Lucas Lehnert, and Michael Littman.
\newblock State abstractions for lifelong reinforcement learning.
\newblock In {\em International Conference on Machine Learning}, pages 10--19.
  PMLR, 2018.

\bibitem{abel2019state}
David Abel, Dilip Arumugam, Kavosh Asadi, Yuu Jinnai, Michael~L Littman, and
  Lawson~LS Wong.
\newblock State abstraction as compression in apprenticeship learning.
\newblock In {\em Proceedings of the AAAI Conference on Artificial
  Intelligence}, volume~33, pages 3134--3142, 2019.

\bibitem{allen2021learning}
Cameron Allen, Neev Parikh, Omer Gottesman, and George Konidaris.
\newblock Learning markov state abstractions for deep reinforcement learning.
\newblock {\em Advances in Neural Information Processing Systems},
  34:8229--8241, 2021.

\bibitem{mnih2015human}
Volodymyr Mnih, Koray Kavukcuoglu, David Silver, Andrei~A Rusu, Joel Veness,
  Marc~G Bellemare, Alex Graves, Martin Riedmiller, Andreas~K Fidjeland, Georg
  Ostrovski, et~al.
\newblock Human-level control through deep reinforcement learning.
\newblock {\em Nature}, 518(7540):529--533, 2015.

\bibitem{alharin2020reinforcement}
Alnour Alharin, Thanh-Nam Doan, and Mina Sartipi.
\newblock Reinforcement learning interpretation methods: A survey.
\newblock {\em IEEE Access}, 8:171058--171077, 2020.

\bibitem{chen2018lifelong}
Zhiyuan Chen and Bing Liu.
\newblock Lifelong machine learning.
\newblock {\em Synthesis Lectures on Artificial Intelligence and Machine
  Learning}, 12(3):1--207, 2018.

\bibitem{khetarpal2020towards}
Khimya Khetarpal, Matthew Riemer, Irina Rish, and Doina Precup.
\newblock Towards continual reinforcement learning: A review and perspectives.
\newblock {\em arXiv preprint arXiv:2012.13490}, 2020.

\bibitem{french1999catastrophic}
Robert~M French.
\newblock Catastrophic forgetting in connectionist networks.
\newblock {\em Trends in cognitive sciences}, 3(4):128--135, 1999.

\bibitem{friston2021sophisticated}
Karl Friston, Lancelot Da~Costa, Danijar Hafner, Casper Hesp, and Thomas Parr.
\newblock Sophisticated inference.
\newblock {\em Neural Computation}, 33(3):713--763, 2021.

\bibitem{blei2006variational}
David~M Blei and Michael~I Jordan.
\newblock Variational inference for {D}irichlet process mixtures.
\newblock {\em Bayesian Analysis}, 1(1):121--143, 2006.

\bibitem{zhang2018advances}
Cheng Zhang, Judith B{\"u}tepage, Hedvig Kjellstr{\"o}m, and Stephan Mandt.
\newblock Advances in variational inference.
\newblock {\em IEEE transactions on pattern analysis and machine intelligence},
  41(8):2008--2026, 2018.

\bibitem{takekawa2009novel}
Takashi Takekawa and Tomoki Fukai.
\newblock A novel view of the variational bayesian clustering.
\newblock {\em Neurocomputing}, 72(13-15):3366--3369, 2009.

\bibitem{sato2001online}
Masa-aki Sato.
\newblock Online model selection based on the variational bayes.
\newblock {\em Neural computation}, 13(7):1649--1681, 2001.

\bibitem{sutton2018reinforcement}
Richard~S Sutton and Andrew~G Barto.
\newblock {\em Reinforcement learning: An introduction}.
\newblock MIT press, 2018.

\bibitem{watkins1992q}
Christopher~JCH Watkins and Peter Dayan.
\newblock {Q}-learning.
\newblock {\em Machine learning}, 8(3-4):279--292, 1992.

\bibitem{agrawal2012analysis}
Shipra Agrawal and Navin Goyal.
\newblock Analysis of {T}hompson sampling for the multi-armed bandit problem.
\newblock In {\em Conference on learning theory}, volume~23, pages 39--1. JMLR
  Workshop and Conference Proceedings, 2012.

\bibitem{cortese2021value}
Aurelio Cortese, Asuka Yamamoto, Maryam Hashemzadeh, Pradyumna Sepulveda,
  Mitsuo Kawato, and Benedetto De~Martino.
\newblock Value signals guide abstraction during learning.
\newblock {\em eLife}, 10:e68943, 2021.

\end{thebibliography}

\end{document}